\documentclass[5p,twocolumn]{elsarticle}

\usepackage{amsmath, amssymb, amsfonts}
\usepackage{microtype}
\usepackage[colorlinks=true, linkcolor=blue!70!black, citecolor=blue!70!black, urlcolor=blue!70!black]{hyperref}
\usepackage{graphicx}
\graphicspath{{figures/}}
\usepackage{booktabs}
\usepackage{multirow}
\usepackage{array}
\usepackage{xcolor}
\usepackage{lineno}

\biboptions{authoryear}

\makeatletter
\def\ps@pprintTitle{%
  \let\@oddhead\@empty
  \let\@evenhead\@empty
  \def\@oddfoot{}%
  \let\@evenfoot\@oddfoot
}
\makeatother

\begin{document}

\begin{frontmatter}

\title{Attention U-Net Cellular Automata for Multi-Class Land Use and Land Cover Change Simulation in Lucknow, India: Evaluation Against Null and Classical Baselines}

\author[vnit]{Shubham Phapale (\href{mailto:shubhamphapale10@gmail.com}{\texttt{shubhamphapale10@gmail.com}})}
\author[vnit]{Poonam Sharma (\href{mailto:poonamsharma@cse.vnit.ac.in}{\texttt{poonamsharma@cse.vnit.ac.in}})}
\address[vnit]{Department of Computer Science and Engineering, Visvesvaraya National Institute of Technology, Nagpur, India}

\begin{abstract}
Convolutional networks are often used to learn the transition rules of land-use cellular automata (CA), yet their predictions are seldom compared with a null model that predicts no change. We built AttentionUNet-CA, which reads a $32 \times 32$ neighbourhood of land-cover states with four drivers (elevation, distance to major roads, distance to the central business district (CBD) and population density) and predicts the state of each cell ten years later. We tested it on Lucknow, India, with GLC\_FCS30D land cover for 1990, 2000, 2010 and 2020, a 30~m series mapped consistently over time, and compared its built-up class with GHSL and ESA WorldCover. AttentionUNet-CA was trained on 1990--2010, selected on held-out spatial blocks and evaluated once on 2010--2020, with three seeds. Baselines were the null model, a random-forest CA (RF-CA) and a U-Net CA without attention; each network was trained with and without class weights. Only 8.1\% of the study area changed in the test decade, so no model matched the overall agreement of the null model (91.9\%). On change, measured by the figure of merit, the class-weighted networks scored highest (U-Net CA 22.0 $\pm$ 1.2\%, AttentionUNet-CA 21.0 $\pm$ 1.4\%), ahead of RF-CA (20.3\%), but by margins comparable to seed-to-seed variation; the networks estimated the amount of change better than RF-CA. The attention gates did not improve accuracy. Population density and distance to the CBD were the most useful drivers. Projected to 2030, built-up land grows by 22--35\%, mostly at the city edge.
\end{abstract}


\end{frontmatter}

\section{Introduction}
\label{sec:intro}

Urban expansion in the Global South is reshaping metropolitan fringes, displacing farmland and putting pressure on regional ecosystems \citep{seto2012global}. In India, much of this growth now happens in secondary cities, and Lucknow, the capital of Uttar Pradesh, is a representative case. The city sits on the alluvial plain of the Gomti River, and over the past three decades it has spread along its radial arterial roads and, more recently, the Amar Shaheed Path bypass. Planning for this growth needs forecasts of where the city will expand, and such forecasts are useful only if their reliability has been tested.

Cellular automata (CA) are the most widely used framework for simulating land-use change \citep{white1993cellular, batty1994cellular}. A CA divides space into a lattice of cells whose states are updated by local transition rules. Classical implementations such as SLEUTH \citep{clarke1997loose} and CA--Markov models combine a Markov estimate of how much change will occur with suitability surfaces from multi-criteria evaluation \citep{eastman1995raster} or logistic regression. FLUS \citep{liu2017flus} and PLUS \citep{liang2021plus} estimate transition potential with machine learning, and deep learning variants go a step further and learn the transition rules directly from neighbourhood images \citep{he2018mining, zhai2020simulating, xing2020novel}. Closer to our setting, \citet{tripathy2019monitoring} used a threshold-based CA with road, central business district (CBD), population density and slope layers to model the growth of Delhi.

Three problems recur in this literature. Many applications still simulate a binary urban/non-urban state, which hides what happens among cropland, vegetation and water. Skill is usually reported as overall accuracy or Kappa over the whole map, and both are dominated by cells that never change. \citet{pontius2008comparing} found that many published land-change models were less accurate than a null model that predicts no change, and argued that models should be judged on the cells that do change; deep learning CA studies rarely make this comparison. Finally, the input maps are often classified separately for each date, so differences between the classifications look like land change, and a model trained on them learns the errors along with the real transitions \citep{olofsson2014good}.

In this paper we build \textit{AttentionUNet-CA}, a convolutional CA that combines a learnable embedding of five land-cover states with four spatial drivers and predicts the next decadal state of a $32 \times 32$ cell neighbourhood with an Attention U-Net \citep{ronneberger2015unet, oktay2018attention}. We evaluate it on Lucknow for 1990--2020 with a multi-class land-cover series (water, vegetation, built-up and agriculture) mapped by one method for every year, and we use the test decade, 2010--2020, only once. The model is compared with a no-change null model, a random-forest CA (RF-CA) \citep{kamusoko2015simulating} and a U-Net CA without attention gates, and each network is trained with and without class weights. All models are scored with the figure of merit (FoM) and its components \citep{pontius2008comparing, varga2019intensity} and with the quantity, exchange and shift components of difference \citep{pontius2014quantity}. Models are selected on spatial blocks \citep{roberts2017cross} held out from the calibration decades (1990--2010), every configuration is run with three seeds, and all metrics are computed inside the study area. We also check the built-up class of the input maps against independent products, look at where the attention gates focus and how much each driver matters, and project the model forward to 2030.

\section{Related work}
\label{sec:related}

\subsection{Cellular automata for land-use change}
Urban CA go back to \citet{tobler1979cellular}. \citet{couclelis1985cellular} argued that realistic urban dynamics need transition rules that vary in space, and \citet{white1993cellular} and \citet{batty1994cellular} showed that fractal urban form can emerge from purely local interactions between cells. In CA--Markov models, the probability of a transition from state $i$ to state $j$ over an interval is estimated by cross-tabulating two maps:
\begin{equation}
    P_{ij} = \frac{n_{ij}}{\sum_{k=1}^{M} n_{ik}}, \qquad \sum_{j=1}^{M} P_{ij} = 1,
    \label{eq:markov_prob}
\end{equation}
where $n_{ij}$ is the number of cells that moved from $i$ to $j$ and $M$ is the number of categories. The Markov matrix fixes how much change occurs; the suitability maps decide where it goes.

\subsection{Machine learning and deep learning in CA}
\citet{li2002neural} used neural networks to estimate multi-class transition potential, and later studies used support vector machines \citep{yang2008urban} and random forests \citep{kamusoko2015simulating}. FLUS \citep{liu2017flus} and PLUS \citep{liang2021plus} combine machine-learned suitability with self-adaptive inertia and patch generation and have become standard benchmarks. All of these describe the neighbourhood of a cell with hand-crafted statistics. Convolutional networks learn it from the data instead: \citet{he2018mining} mined CA transition rules with a CNN, \citet{zhai2020simulating} combined a CNN with a vector-based CA, and \citet{xing2020novel} coupled a CA with recurrent and convolutional networks. U-Nets \citep{ronneberger2015unet} keep spatial detail through skip connections, and the Attention U-Net \citep{oktay2018attention} places additive attention gates on those connections. \citet{wang2022unet} used a U-Net to simulate urban land-use patterns and their dynamics across the North China Plain, but attention-gated U-Nets have received little evaluation for multi-class, decade-long simulation of land use and land cover.

\subsection{Validation of land-change models}
\citet{pontius2011extreme} recommended replacing Kappa with quantity and allocation disagreement, and \citet{pontius2014quantity} later split allocation disagreement into exchange and shift. Since most cells persist between two dates, overall agreement mostly measures persistence. For that reason \citet{pontius2008comparing} compared models against a null model of no change and used the figure of merit, the ratio of correctly simulated change to the union of observed and simulated change. \citet{varga2019intensity} later recommended reporting its components (misses, hits, wrong hits and false alarms) as well. We use all of these measures here.

\section{Study area and data}
\label{sec:data}

\subsection{Lucknow}
Lucknow ($26^\circ 51'$N, $80^\circ 57'$E) lies in the central Gangetic plain; in the Copernicus DEM \citep{esa2022copdem}, the study area has an average elevation of about 120~m. The Gomti River runs north-west to south-east through the city and separates the older Cis-Gomti areas from the Trans-Gomti suburbs. The city has grown along the radial roads to Sitapur, Kanpur, Faizabad (Ayodhya) and Sultanpur and, more recently, along the Amar Shaheed Path bypass in the south. Our study area covers 1,223~km$^2$ of the city and its peri-urban fringe. According to GHS-POP \citep{schiavina2023ghspop}, its population grew from about 2.4 million in 1990 to 4.1 million in 2010 and 4.9 million in 2020.

\subsection{Land use and land cover maps}
We took land cover from GLC\_FCS30D \citep{zhang2024glcfcs30d}, a global 30~m product for 1985--2022 derived from dense Landsat time series with a continuous change-detection method. Every year is mapped the same way, so the series is consistent over time, which is important when transition rules are learned from it. Its authors report an overall accuracy of 80.9\% for ten basic land-cover types. We used the maps for 1990, 2000, 2010 and 2020 and merged the GLC\_FCS30D classes into four categories: water (open water and wetlands), vegetation (forest, shrubland, grassland and sparse vegetation), built-up (impervious surfaces) and agriculture (rainfed and irrigated cropland, with bare and fallow land). The 30~m maps were resampled by majority to a grid of $959 \times 1221$ cells of $0.00049^\circ$ (about $49 \times 54$~m). Of these cells, 464,441 fall inside the study area and the rest are coded as background. Table~\ref{tab:class_distribution} lists the area of each category.

\begin{table*}[t]
\centering
\caption{Area of each category within the study area, in cells (percentage of the study area in parentheses).}
\label{tab:class_distribution}
\begin{tabular}{lrrrr}
\toprule
Category & 1990 & 2000 & 2010 & 2020 \\
\midrule
Water & 1,082 (0.23) & 1,199 (0.26) & 1,366 (0.29) & 1,234 (0.27) \\
Vegetation & 16,082 (3.46) & 14,950 (3.22) & 14,607 (3.15) & 13,707 (2.95) \\
Built-up & 40,309 (8.68) & 57,892 (12.46) & 101,929 (21.95) & 122,862 (26.45) \\
Agriculture & 406,968 (87.63) & 390,400 (84.06) & 346,539 (74.61) & 326,638 (70.33) \\
\midrule
Total & 464,441 & 464,441 & 464,441 & 464,441 \\
\bottomrule
\end{tabular}
\end{table*}

\subsection{Independent check of the maps}
\label{sec:check}
We had no field reference data, so we checked the maps against two independent products instead. GHS-BUILT-S, the built-up surface grid of the Global Human Settlement Layer (GHSL) \citep{pesaresi2023ghsbuilt}, covers the same four years. Cell by cell, it agrees with our built-up class with an F1 of 70.6\% in 1990, rising to 79.7\% in 2020 (Table~\ref{tab:data_check}). Both products show built-up land growing in every decade, but GLC\_FCS30D maps more of it, especially after 2000. One likely reason is that its 30~m impervious class picks up roads and mixed peri-urban pixels that fall below the 20\% built-up threshold we used for GHSL.

\begin{table}[t]
\centering
\footnotesize
\caption{Built-up land in the GLC\_FCS30D maps compared with GHS-BUILT-S, counting a GHSL cell as built-up when at least 20\% of it is built. F1 measures the cell-by-cell agreement within the study area.}
\label{tab:data_check}
\setlength{\tabcolsep}{4pt}
\begin{tabular}{lrrr}
\toprule
Year & GLC\_FCS30D (km$^2$) & GHSL (km$^2$) & F1 (\%) \\
\midrule
1990 & 106 & 95 & 70.6 \\
2000 & 152 & 133 & 74.8 \\
2010 & 268 & 181 & 75.6 \\
2020 & 324 & 255 & 79.7 \\
\bottomrule
\end{tabular}
\end{table}

ESA WorldCover 2021 \citep{zanaga2022worldcover} allows all four categories to be compared. The 2020 map and WorldCover agree on 62.8\% of the study area: well for built-up land (F1 84.7\%) and agriculture (70.9\%), poorly for vegetation (12.1\%). WorldCover puts 36.0\% of the study area under trees, shrubs or grass, whereas only 3.0\% is vegetation in GLC\_FCS30D. Most of the difference is probably trees around villages and in orchards, and grass on fallow land, all of which GLC\_FCS30D labels as cropland. Vegetation is therefore under-represented in our maps, and results for this category need to be read with care. Built-up land, where most of the change happens, is well reproduced.

\subsection{Spatial driver layers}
We prepared four drivers on the same grid. Elevation is the Copernicus DEM GLO-30 \citep{esa2022copdem} averaged to the grid; within the study area it varies only from 101 to 142~m. Distance to major roads (road distance) is the Euclidean distance to the nearest motorway, trunk, primary or secondary road in OpenStreetMap \citep{osm2026}. Distance to the central business district (CBD) is measured from Hazratganj. Population density comes from GHS-POP \citep{schiavina2023ghspop}. Each decade uses the layer for its first year (1990, 2000 or 2010 for the observed decades and 2020 for the projection), so the population layer never postdates the start of the decade being simulated. GHS-POP puts people only on built-up pixels and leaves most farmland at zero, so we smoothed it with a Gaussian kernel ($\sigma = 500$~m) into a regional density surface and took its logarithm. Each layer was then scaled to $[0,1]$ inside the study area; the four population layers share one scale, so that growth between the years is preserved.

\subsection{Observed transitions}
\label{sec:transitions}
Table~\ref{tab:transitions} gives the row-normalised transition matrices (Eq.~\ref{eq:markov_prob}) for the three decades.

\begin{table*}[t]
\centering
\caption{Row-normalised transition matrices (\%) within the study area. Rows give the state at the start of the decade and columns the state at the end. Diagonal entries are persistence.}
\label{tab:transitions}
\begin{tabular}{llrrrr}
\toprule
Decade & From \textbackslash\ To & Water & Vegetation & Built-up & Agriculture \\
\midrule
\multirow{4}{*}{1990$\to$2000} & Water & 83.5 & 1.0 & 0.6 & 14.9 \\
& Vegetation & 0.3 & 70.4 & 6.4 & 22.9 \\
& Built-up & 0.0 & 0.1 & 98.1 & 1.8 \\
& Agriculture & 0.1 & 0.9 & 4.3 & 94.8 \\
\midrule
\multirow{4}{*}{2000$\to$2010} & Water & 89.7 & 2.0 & 1.4 & 6.9 \\
& Vegetation & 0.1 & 72.9 & 5.2 & 21.8 \\
& Built-up & 0.0 & 0.2 & 98.6 & 1.1 \\
& Agriculture & 0.1 & 0.9 & 11.3 & 87.7 \\
\midrule
\multirow{4}{*}{2010$\to$2020} & Water & 78.3 & 2.0 & 1.5 & 18.2 \\
& Vegetation & 0.2 & 61.3 & 13.7 & 24.7 \\
& Built-up & 0.0 & 0.5 & 95.9 & 3.6 \\
& Agriculture & 0.0 & 1.2 & 6.7 & 92.1 \\
\bottomrule
\end{tabular}
\end{table*}

Change was moderate and consisted mainly of urban growth: 5.8\% of the study area changed category in 1990--2000, 11.4\% in 2000--2010 and 8.1\% in 2010--2020. Built-up land went from 8.7\% of the area in 1990 to 26.5\% in 2020 (from 106 to 324~km$^2$), and nearly all of the growth took place on former farmland. Agriculture lost 4.3\%, 11.3\% and 6.7\% of its area to built-up land in the three decades. Built-up land was highly persistent (95.9--98.6\%). In the test decade two-thirds (66.8\%) of all changed cells became built-up, and 89.3\% of the new built-up cells were within 200~m of land already built in 2010, so the city grew mostly outward from its edge and along its roads. The few reverse transitions, such as the 3.6\% of built-up cells shown as agriculture in 2020, are more likely to be classification errors than real change. Water and vegetation cover little ground but change substantially in relative terms; 18.2\% of the 2010 water cells, for instance, were agriculture by 2020, which probably reflects the shifting margins of rivers and ponds.

\section{Methods}
\label{sec:methods}

\subsection{Problem formulation}
Let $\Lambda$ be the $H \times W$ lattice ($H = 959$, $W = 1221$). At time $t$ each cell $(x,y)$ has a state $L_t(x,y) \in \{0,1,2,3,4\}$ (water, vegetation, built-up, agriculture, background) and a driver vector $\mathbf{d}_t(x,y) \in [0,1]^4$. A CA transition rule maps the state of a cell, its neighbourhood $\mathcal{N}(x,y)$ and its drivers to the next state:
\begin{equation}
    L_{t+1}(x,y) = f\big(L_{t}(x,y), \mathcal{N}(x,y), \mathbf{d}_t(x,y)\big).
\end{equation}
We replace $f$ with a network $f_\theta$ that takes a $32 \times 32$ window of states and drivers and returns class probabilities for every cell in the window. Background cells lie outside the study area and are never changed.

\subsection{AttentionUNet-CA}
The architecture has 8,661,758 trainable parameters.

\paragraph{Input} The state window $\mathbf{L}_t \in \{0,\dots,4\}^{32 \times 32}$ is passed through a learnable embedding of dimension $d_e = 16$, which is equivalent to a linear projection of a one-hot encoding. The embedded states are concatenated with the four driver channels to form $\mathbf{X}_0 \in \mathbb{R}^{20 \times 32 \times 32}$.

\paragraph{Encoder} Three stages of a double convolution block
\begin{equation}
    \mathcal{C}(z) = \text{ReLU}\big(\text{BN}(\text{Conv}_{3\times3}(\text{ReLU}(\text{BN}(\text{Conv}_{3\times3}(z)))))\big)
\end{equation}
with $2 \times 2$ max pooling produce feature maps with 64, 128 and 256 channels at resolutions 32, 16 and 8. A bottleneck block applied after a further pooling step yields 512 channels at $4 \times 4$.

\paragraph{Attention-gated decoder} Each decoder stage upsamples by nearest-neighbour interpolation followed by a $3 \times 3$ convolution, batch normalisation and ReLU, giving a gating signal $\mathbf{g}$ at the resolution of the corresponding encoder features $\mathbf{x}$. An additive attention gate \citep{oktay2018attention} computes
\begin{align}
    \boldsymbol{\psi} &= \text{ReLU}\big(\text{BN}(\mathbf{W}_g * \mathbf{g}) + \text{BN}(\mathbf{W}_x * \mathbf{x})\big), \\
    \boldsymbol{\alpha} &= \sigma\big(\text{BN}(\mathbf{W}_\psi * \boldsymbol{\psi})\big) \in [0,1]^{H' \times W'}, \qquad \hat{\mathbf{x}} = \boldsymbol{\alpha} \odot \mathbf{x},
\end{align}
where $\mathbf{W}_g$, $\mathbf{W}_x$ and $\mathbf{W}_\psi$ are $1 \times 1$ convolutions. The gated skip features $\hat{\mathbf{x}}$ are concatenated with $\mathbf{g}$ and passed through a convolution block. A final $1 \times 1$ convolution produces five logits per cell. The U-Net CA baseline is identical except that the skip features are concatenated without gating (8,574,165 parameters).

\subsection{Training and model selection}
Models are trained on the two calibration decades, 1990$\to$2000 and 2000$\to$2010. The 2010$\to$2020 decade is kept for testing; it was not used for training, model selection or any design decision before the models were evaluated on it. For model selection we divide the grid into $128 \times 128$ cell blocks and hold out a random 20\% of the blocks that lie mostly inside the study area \citep{roberts2017cross}. From both calibration decades we extract $32 \times 32$ patches with a stride of 16. Patches entirely inside training blocks form the training set, patches entirely inside held-out blocks form the validation set, and patches that cross a block boundary or have less than 10\% study-area cells are dropped. This leaves 3,128 training and 604 validation patches.

The loss is the cross-entropy over study-area cells, with background cells ignored:
\begin{equation}
    \mathcal{L}(\theta) = -\frac{\sum_{n} w_{y_n} \log \hat{y}_{n, y_n}}{\sum_{n} w_{y_n}},
    \label{eq:loss}
\end{equation}
where the sum runs over all non-background target cells in a batch, $y_n$ is the observed class of cell $n$ and $\hat{y}_{n,c}$ the predicted probability of class $c$. The unweighted model uses $w_c = 1$. The class-weighted model uses smoothed inverse-frequency weights
\begin{equation}
    w_c = 4\,\frac{f_c^{-1/2}}{\sum_{k=0}^{3} f_k^{-1/2}},
\end{equation}
where $f_c$ is the frequency of class $c$ among study-area cells of the 2000 and 2010 target maps. The resulting weights are $\mathbf{w} = (2.70, 0.80, 0.34, 0.16)$ for water, vegetation, built-up and agriculture.

We train with Adam \citep{kingma2015adam} at a learning rate of $10^{-3}$, with batch size 32, gradient clipping at norm 1 and 20 epochs. The learning rate is halved when the validation macro-averaged F1 has not improved for two epochs, and we keep the epoch with the best validation macro F1. Each configuration is trained with three random seeds (42, 43, 44), and we report the mean and standard deviation.

\subsection{Full-lattice simulation}
To simulate a full map, we slide the $32 \times 32$ window with a stride of 16 and blend the overlapping class probabilities with a two-dimensional Hanning window, so that every cell is predicted with context on all sides. This was decided before any testing. Each cell then takes the class with the highest blended probability among the four study-area classes.

\subsection{Baselines}
\paragraph{Null model} The 2010 map is used unchanged as the simulation of 2020 \citep{pontius2008comparing}.
\paragraph{RF-CA} A random forest \citep{kamusoko2015simulating} with 100 trees (maximum depth 16, balanced class weights) predicts the next state of each cell from 14 features: the one-hot current state, the four drivers and the class composition of the $3 \times 3$ Moore neighbourhood. Up to 35,000 cells per class and decade are sampled from the training blocks.
\paragraph{U-Net CA} The AttentionUNet-CA without attention gates, trained with and without class weights.

\subsection{Evaluation}
All metrics are computed over the 464,441 study-area cells of the 2020 reference map. We report overall agreement (OA); quantity, exchange and shift disagreement, whose sum equals $1 - \text{OA}$ \citep{pontius2014quantity}; and per-class precision, recall and F1. Relative to the 2010 map, each cell falls into one of five outcomes: correct persistence, misses (observed change simulated as persistence), hits (observed change simulated correctly), wrong hits (observed change simulated as change to the wrong class) and false alarms (observed persistence simulated as change). The figure of merit is
\begin{equation}
    \text{FoM} = \frac{\text{hits}}{\text{misses} + \text{hits} + \text{wrong hits} + \text{false alarms}}.
\end{equation}
Cohen's Kappa is not reported, following \citet{pontius2011extreme}.

\section{Results}
\label{sec:results}

\subsection{Agreement with the 2020 map and comparison with the null model}
Table~\ref{tab:results_main} summarises the simulation of 2020 from 2010. Because only 8.08\% of the study area changed, the null model already agrees with 2020 on 91.92\% of it, and none of the models does better. The unweighted networks come closest (91.71--91.85\%), the class-weighted networks reach 89.51--89.98\%, and RF-CA is lowest at 85.20\%. This is the same pattern \citet{pontius2008comparing} found for most of the models they compared, and it is why overall agreement on its own says little about simulation skill.

The figure of merit looks only at change. Here the class-weighted U-Net CA scores highest (22.00 $\pm$ 1.18\%), followed by the class-weighted AttentionUNet-CA (20.96 $\pm$ 1.39\%), the unweighted U-Net CA (20.75 $\pm$ 0.42\%) and RF-CA (20.32 $\pm$ 0.04\%). These margins are comparable to the variation between seeds. The unweighted AttentionUNet-CA is lowest at 19.27 $\pm$ 0.43\%. The networks are much better than RF-CA at getting the amount of change right: their quantity disagreement is 0.7--3.9\%, compared with 10.1\% for RF-CA and 4.5\% for the null model.

\begin{table*}[t]
\centering
\caption{Simulation of 2020 from 2010 within the study area (464,441 cells). Values are percentages; network and RF-CA values give the mean $\pm$ standard deviation over three seeds. Quantity, exchange and shift disagreement sum to $100 - \text{OA}$.}
\label{tab:results_main}
\small
\setlength{\tabcolsep}{5pt}
\begin{tabular}{lrrrrrr}
\toprule
Model & OA & Quantity & Exchange & Shift & FoM & Macro F1 \\
\midrule
No-change (null) & 91.92 & 4.51 & 3.41 & 0.16 & 0.00 & 81.83 \\
RF-CA & 85.20 $\pm$ 0.05 & 10.13 $\pm$ 0.08 & 4.55 $\pm$ 0.04 & 0.13 $\pm$ 0.01 & 20.32 $\pm$ 0.04 & 70.36 $\pm$ 0.17 \\
U-Net CA, unweighted & 91.71 $\pm$ 0.16 & 0.74 $\pm$ 0.30 & 7.36 $\pm$ 0.50 & 0.18 $\pm$ 0.04 & 20.75 $\pm$ 0.42 & 81.71 $\pm$ 0.21 \\
U-Net CA, class-weighted & 89.51 $\pm$ 0.38 & 3.89 $\pm$ 0.94 & 6.42 $\pm$ 0.57 & 0.18 $\pm$ 0.06 & 22.00 $\pm$ 1.18 & 76.54 $\pm$ 0.36 \\
AttentionUNet-CA, unweighted & 91.85 $\pm$ 0.12 & 1.20 $\pm$ 0.29 & 6.75 $\pm$ 0.40 & 0.19 $\pm$ 0.03 & 19.27 $\pm$ 0.43 & 81.77 $\pm$ 0.26 \\
AttentionUNet-CA, class-weighted & 89.98 $\pm$ 0.22 & 2.78 $\pm$ 0.78 & 7.16 $\pm$ 0.59 & 0.09 $\pm$ 0.06 & 20.96 $\pm$ 1.39 & 77.21 $\pm$ 0.76 \\
\bottomrule
\end{tabular}
\end{table*}

\subsection{Components of change}
\begin{figure}[t]
\centering
\includegraphics[width=\columnwidth]{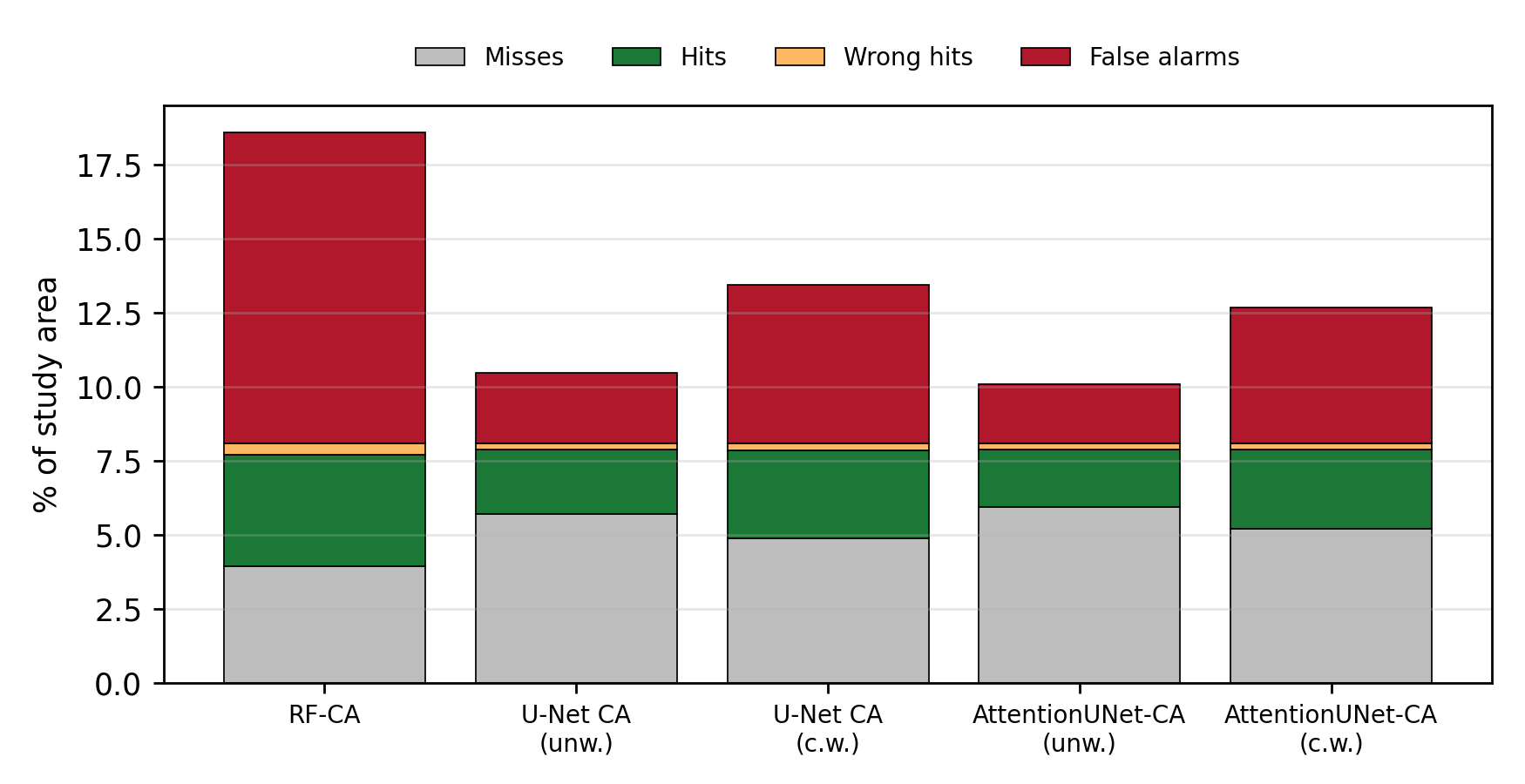}
\caption{Components of the figure of merit for each model (mean of three seeds), as percentages of the study area (unw.: unweighted; c.w.: class-weighted).}
\label{fig:fom}
\end{figure}

Table~\ref{tab:fom_components} and Fig.~\ref{fig:fom} split the FoM into its parts. RF-CA simulates far more change than actually happened, 14.64\% of the study area against 8.08\% observed. That gives it the most hits (3.77\%), but also by a wide margin the most false alarms (10.50\%). The class-weighted AttentionUNet-CA simulates 7.46\% change, close to the observed amount, with 2.66\% hits and 4.61\% false alarms; about a third (36\%) of the cells it changes are hits, against a quarter (26\%) for RF-CA. The class-weighted U-Net CA is similar, with 8.57\% simulated change and 2.96\% hits. The unweighted networks are more cautious still. They change only 4.2--4.8\% of the area and so miss most of what happened, but when they do predict change they are right more often (46--47\% hits).

\begin{table*}[t]
\centering
\caption{Components of the figure of merit, as percentages of the study area, and the share of the study area simulated as changing. The observed change between 2010 and 2020 is 8.08\%.}
\label{tab:fom_components}
\begin{tabular}{lrrrrr}
\toprule
Model & Misses & Hits & Wrong hits & False alarms & Simulated change \\
\midrule
No-change (null) & 8.08 & 0.00 & 0.00 & 0.00 & 0.00 \\
RF-CA & 3.94 $\pm$ 0.02 & 3.77 $\pm$ 0.02 & 0.37 $\pm$ 0.01 & 10.50 $\pm$ 0.06 & 14.64 $\pm$ 0.08 \\
U-Net CA, unweighted & 5.71 $\pm$ 0.09 & 2.17 $\pm$ 0.09 & 0.20 $\pm$ 0.00 & 2.38 $\pm$ 0.25 & 4.75 $\pm$ 0.34 \\
U-Net CA, class-weighted & 4.88 $\pm$ 0.33 & 2.96 $\pm$ 0.30 & 0.24 $\pm$ 0.04 & 5.37 $\pm$ 0.66 & 8.57 $\pm$ 0.98 \\
AttentionUNet-CA, unweighted & 5.95 $\pm$ 0.07 & 1.95 $\pm$ 0.08 & 0.19 $\pm$ 0.00 & 2.02 $\pm$ 0.20 & 4.15 $\pm$ 0.27 \\
AttentionUNet-CA, class-weighted & 5.22 $\pm$ 0.28 & 2.66 $\pm$ 0.26 & 0.20 $\pm$ 0.03 & 4.61 $\pm$ 0.46 & 7.46 $\pm$ 0.73 \\
\bottomrule
\end{tabular}
\end{table*}

\begin{figure*}[t]
\centering
\includegraphics[width=\textwidth]{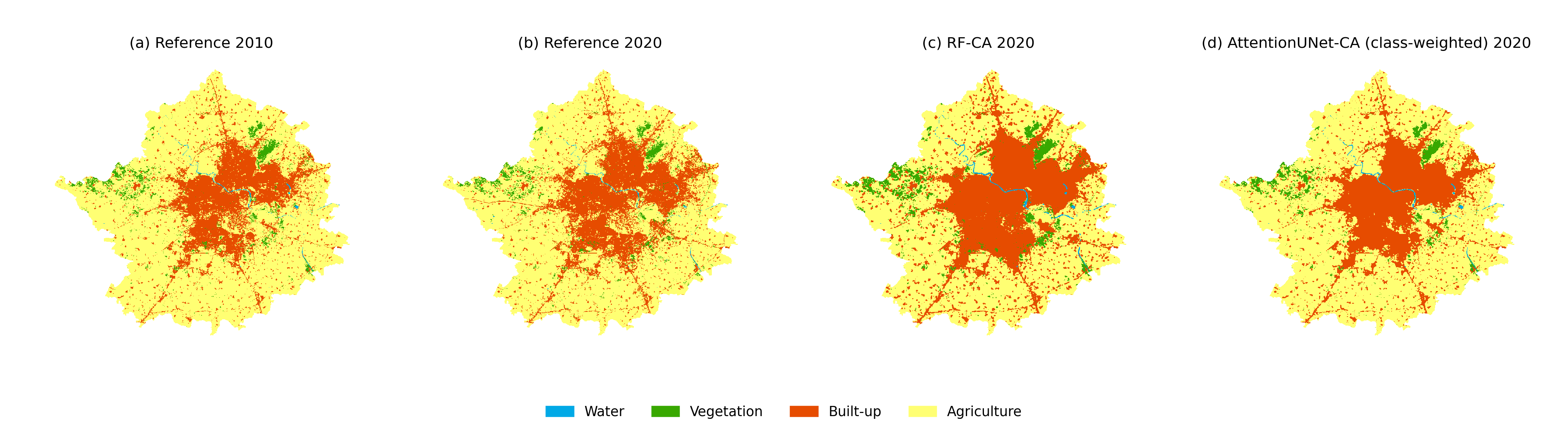}
\caption{Observed land cover in 2010 and 2020 and the 2020 simulations of RF-CA and the class-weighted AttentionUNet-CA (seed 42).}
\label{fig:maps}
\end{figure*}

Most of the observed change is new built-up land. Of the 25,077 cells built up between 2010 and 2020, the class-weighted AttentionUNet-CA places 12,017 (48\%) correctly and RF-CA 16,413 (65\%). RF-CA achieves this by converting 57,343 cells, 2.3 times the number actually built, whereas AttentionUNet-CA converts 32,270 (1.3 times). In Fig.~\ref{fig:maps}, both models extend the 2010 built-up area outward and along the main roads, and both make the city more compact than the reference map by filling in open land between built-up patches. None of them removes built-up land; every model keeps more than 99.9\% of the 2010 built-up cells.

\subsection{Per-category performance}
Table~\ref{tab:per_class} gives recall and F1 by category. Agriculture and built-up land are simulated best. For built-up land the null model already reaches an F1 of 87.0\%, since most of the built-up cells of 2020 were already built in 2010; the networks reach 85.0--87.4\% and RF-CA 80.9\%. Class weighting lifts built-up recall from 86--87\% to 89--90\%, but at some cost in precision, as the class-weighted models also convert cells that stayed agricultural. Water and vegetation, about 3\% of the study area between them, are harder. For both, the class-weighted models and RF-CA have a lower F1 than the null model, because the extra weight on rare classes produces false conversions into them.

\begin{table*}[t]
\centering
\footnotesize
\caption{Per-category recall (R) and F1 (\%) for the 2020 simulation.}
\label{tab:per_class}
\setlength{\tabcolsep}{3pt}
\begin{tabular}{lrrrrrrrr}
\toprule
& \multicolumn{2}{c}{Water} & \multicolumn{2}{c}{Vegetation} & \multicolumn{2}{c}{Built-up} & \multicolumn{2}{c}{Agriculture} \\
\cmidrule(lr){2-3}\cmidrule(lr){4-5}\cmidrule(lr){6-7}\cmidrule(lr){8-9}
Model & R & F1 & R & F1 & R & F1 & R & F1 \\
\midrule
No-change (null) & 86.6 & 82.2 & 65.4 & 63.3 & 79.6 & 87.0 & 97.7 & 94.8 \\
RF-CA & 92.1 $\pm$ 0.3 & 57.9 $\pm$ 0.5 & 71.3 $\pm$ 0.2 & 53.3 $\pm$ 0.1 & 92.9 $\pm$ 0.1 & 80.9 $\pm$ 0.1 & 82.9 $\pm$ 0.1 & 89.3 $\pm$ 0.0 \\
U-Net CA, unweighted & 85.7 $\pm$ 0.6 & 80.3 $\pm$ 0.4 & 58.8 $\pm$ 0.8 & 64.8 $\pm$ 0.2 & 87.1 $\pm$ 0.4 & 87.3 $\pm$ 0.2 & 94.8 $\pm$ 0.4 & 94.4 $\pm$ 0.1 \\
U-Net CA, class-weighted & 91.0 $\pm$ 0.2 & 66.7 $\pm$ 1.2 & 64.7 $\pm$ 1.5 & 61.7 $\pm$ 0.5 & 90.4 $\pm$ 1.1 & 85.0 $\pm$ 0.5 & 90.2 $\pm$ 0.9 & 92.8 $\pm$ 0.3 \\
AttentionUNet-CA, unweighted & 86.1 $\pm$ 1.2 & 80.2 $\pm$ 0.8 & 59.1 $\pm$ 0.4 & 64.9 $\pm$ 0.1 & 86.3 $\pm$ 0.3 & 87.4 $\pm$ 0.2 & 95.3 $\pm$ 0.3 & 94.5 $\pm$ 0.1 \\
AttentionUNet-CA, class-weighted & 91.1 $\pm$ 0.8 & 67.9 $\pm$ 3.5 & 64.1 $\pm$ 0.6 & 62.4 $\pm$ 0.9 & 89.3 $\pm$ 1.0 & 85.4 $\pm$ 0.3 & 91.3 $\pm$ 0.7 & 93.1 $\pm$ 0.2 \\
\bottomrule
\end{tabular}
\end{table*}

\subsection{Effect of class weighting and of attention gates}
\label{sec:weighting}
Class weighting affects the amount of simulated change far more than the architecture does. With weighting, both networks simulate about 1.8 times as much change (7.5--8.6\% instead of 4.2--4.8\%), and FoM rises by 1.3 points for U-Net CA and 1.7 points for AttentionUNet-CA, gains of about one standard deviation between seeds. The price is more false alarms, which lowers overall agreement by 1.9--2.2 points and macro F1 by about 5 points.

The attention gates, on the other hand, do not improve accuracy. With weighting, U-Net CA has the higher FoM by 1.0 point, which is within the seed-to-seed variation (standard deviation 1.18--1.39 points), and AttentionUNet-CA the higher OA by 0.5 points. Without weighting, U-Net CA again has the higher FoM (20.75\% against 19.27\%), and here the gap is larger than the seed-to-seed variation.

Table~\ref{tab:validation} sets validation against test performance. On the held-out blocks of the calibration decades the unweighted models have the higher macro F1 (80.3--80.4\% against 76.0--76.5\%), and the same holds on the test decade (81.7--81.8\% against 76.5--77.2\%). Selecting by validation macro F1 would therefore have picked the unweighted AttentionUNet-CA, which simulates only about half of the observed change and has the lowest FoM of the four networks. The advantage of the class-weighted models shows up only in FoM; macro F1 over all cells, most of which never change, hides it. This is why we report all four configurations.

\begin{table}[t]
\centering
\footnotesize
\caption{Macro F1 on the held-out blocks of the calibration decades (mean of three seeds) compared with macro F1 and FoM on the 2010$\to$2020 test decade (unw.: unweighted; c.w.: class-weighted).}
\label{tab:validation}
\setlength{\tabcolsep}{3pt}
\begin{tabular}{lrrr}
\toprule
Model & Val.\ F1 & Test F1 & Test FoM \\
\midrule
U-Net CA, unw. & 80.33 & 81.71 $\pm$ 0.21 & 20.75 $\pm$ 0.42 \\
U-Net CA, c.w. & 76.03 & 76.54 $\pm$ 0.36 & 22.00 $\pm$ 1.18 \\
AttentionUNet-CA, unw. & 80.44 & 81.77 $\pm$ 0.26 & 19.27 $\pm$ 0.43 \\
AttentionUNet-CA, c.w. & 76.54 & 77.21 $\pm$ 0.76 & 20.96 $\pm$ 1.39 \\
\bottomrule
\end{tabular}
\end{table}

\subsection{Sensitivity to the driver layers}
\begin{figure}[t]
\centering
\includegraphics[width=\columnwidth]{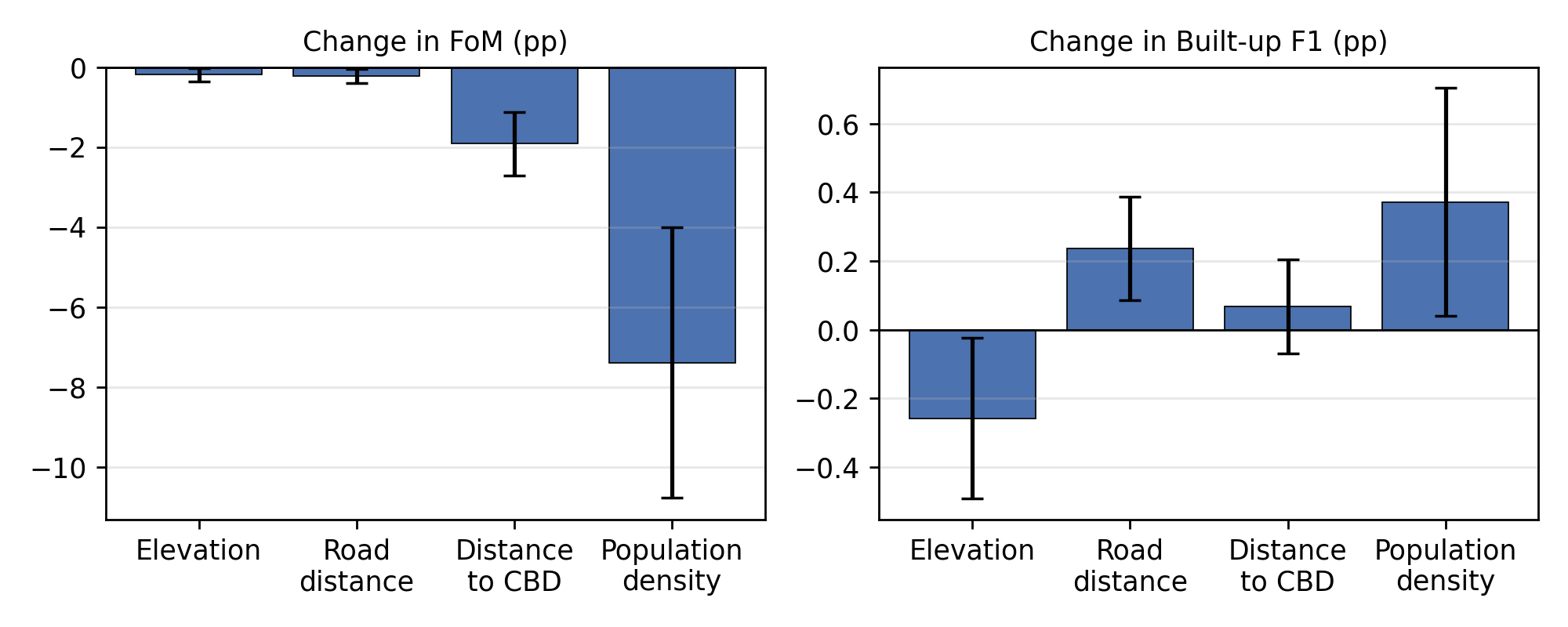}
\caption{Change in FoM and built-up F1 when one driver is permuted within the study area (class-weighted AttentionUNet-CA; mean $\pm$ s.d.\ over 3 seeds $\times$ 5 permutations).}
\label{fig:drivers}
\end{figure}

To see how much the class-weighted AttentionUNet-CA relies on each driver, we shuffled the values of one driver among the study-area cells, which destroys its spatial pattern but keeps its distribution, and simulated 2020 again. We did this five times per driver for each of the three seeds. Fig.~\ref{fig:drivers} shows the mean change in FoM and built-up F1.

Population density matters most. Shuffling it cut FoM by 7.4 points on average (3.4--11.4 across seeds). Shuffling distance to the CBD cut it by 1.9 points (0.8--2.5). Shuffling road distance or elevation changed FoM by only 0.2 points, so the model gains little from either; for elevation this is consistent with a range of only about 40~m across the study area. Population density and distance to the CBD ranked first and second for all three seeds. Built-up F1 changed by less than one point whichever driver was shuffled, so the drivers inform the model mainly about where change is likely rather than about which cells remain built-up.

\subsection{Attention coefficients}
We also recorded the attention coefficients $\boldsymbol{\alpha}$ of the finest gate over the whole lattice for the class-weighted AttentionUNet-CA. Averaged over the seeds, $\boldsymbol{\alpha}$ was 0.63 at cells that changed between 2010 and 2020 and 0.52 at cells that did not, and every seed showed the same gap (0.62--0.65 against 0.48--0.54). Over water the coefficients were close to 1 (0.97), and at cells that became built-up they averaged 0.59. The gates therefore pass more of the skip-connection features at changing cells, but this does not improve accuracy: the U-Net CA uses the same features without gating and performs at least as well.

\subsection{Projection to 2030}
\label{sec:projection}
To show how the model could be used, we retrained the class-weighted AttentionUNet-CA on all three decades (1990--2000, 2000--2010 and 2010--2020, each with the population layer of its first year; 20 epochs, three seeds, the same validation blocks) and ran it once on the 2020 map with the 2020 population layer. We used this configuration because it is the model proposed in this paper; its FoM on the test decade was within the seed-to-seed variation of the best configuration, the class-weighted U-Net CA. On the validation blocks alone, the unweighted AttentionUNet-CA would have been chosen (Section~\ref{sec:weighting}). There is no 2030 map to check against, so the result is a continuation of past transitions, not a forecast.

\begin{figure*}[t]
\centering
\includegraphics[width=\textwidth]{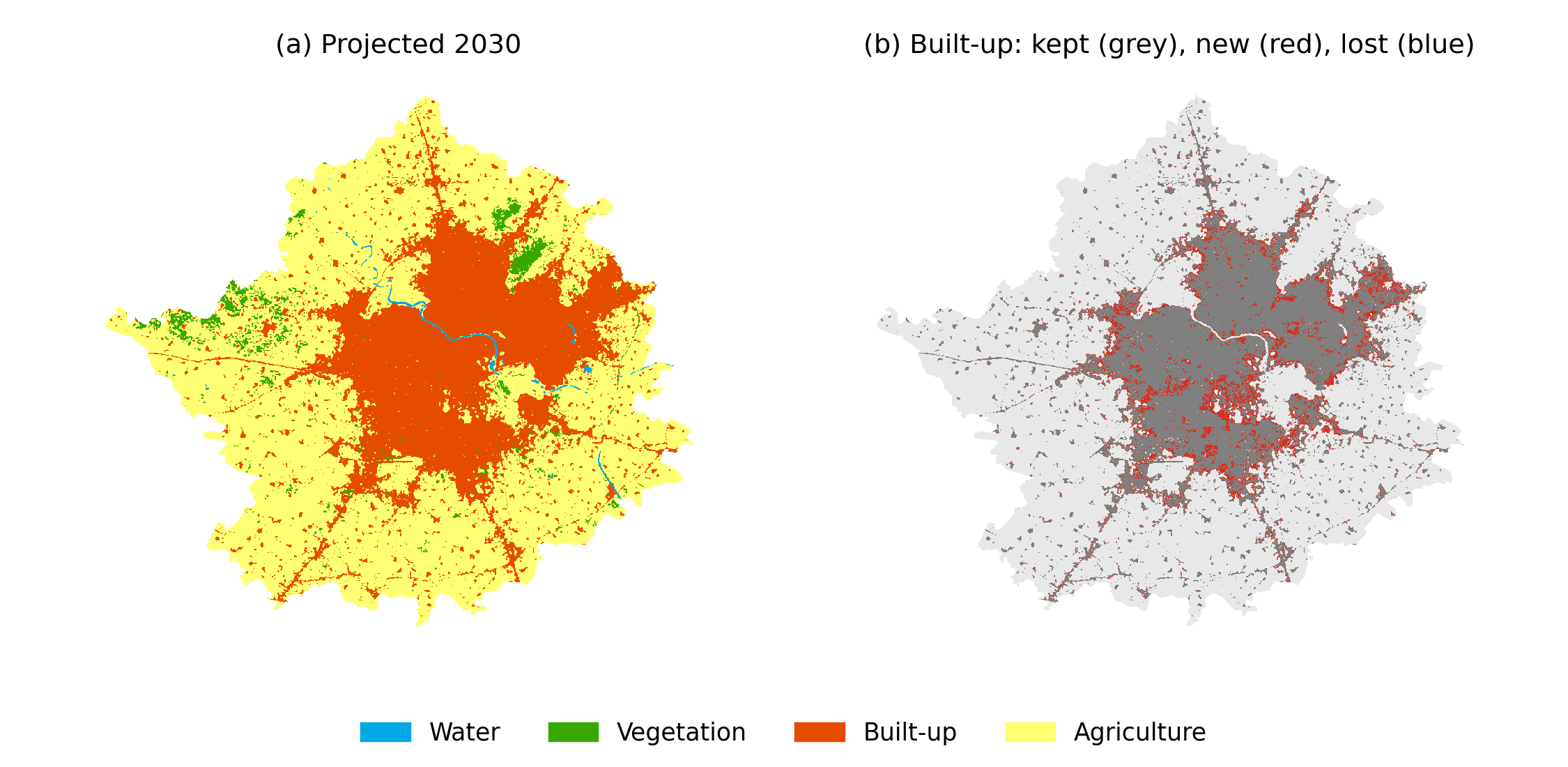}
\caption{Projection to 2030 by the class-weighted AttentionUNet-CA trained on all three decades (seed 42). (a) Projected land use. (b) Built-up cells kept from 2020 (grey), newly built-up (red) and leaving the built-up class (blue).}
\label{fig:projection}
\end{figure*}

The three seeds agree on 95.3\% of the study area but differ in how much land they convert. They put built-up land at 32.2--35.7\% of the study area in 2030, up from 26.5\% in 2020, an increase of 21.7--35.0\% or roughly 70--113~km$^2$. That is at or above the pace of the last observed decade, when built-up land grew by 20.5\%. Nearly all of the new built-up cells (98.5--99.5\%) lie within 200~m of the 2020 built-up area, at the city edge and along the radial roads (Fig.~\ref{fig:projection}), and almost no built-up land is lost. Agriculture shrinks from 70.3\% to 61.1--64.7\% of the study area. The projected city grows more compactly than the real one did. In 2010--2020, 89.3\% of the new built-up cells were within 200~m of existing built-up land, so some real growth took place farther out, and the model does not create settlements of that kind.

\section{Discussion}
\label{sec:discussion}

\subsection{What the models learn}
All models capture the main process in Lucknow, the expansion of the built-up area into the surrounding farmland, and all of them treat built-up land as permanent. The class-weighted networks simulate roughly the right amount of change and put 48--53\% of the new built-up land in the right place. RF-CA finds more of the new built-up land (65\%), but only by converting more than twice as many cells as were actually built. For planning applications, where the total amount of new urban land matters as much as its location, the smaller quantity disagreement of the networks is a practical advantage.

Still, the networks miss 60--74\% of the observed change and RF-CA about half. In the seed-42 runs of AttentionUNet-CA and RF-CA, 44--56\% of the missed cells are changes among agriculture, vegetation and water, which the models barely simulate at all; the rest is new built-up land. More than 98\% of the correctly simulated new built-up cells lie within 200~m of existing built-up land, but 18--29\% of the missed ones lie farther away, so growth that jumps away from the city is not anticipated. Within the urban fringe, the models also cannot distinguish which of many similar cells will be developed within a decade.

\subsection{Evaluation practice}
Two lessons from the land-change literature hold here. Overall agreement and Kappa are not enough on their own \citep{pontius2011extreme}: in this study the null model has the highest agreement, and among the networks the ones with the highest agreement are those that simulate the least change. It is the comparison with a null model, together with the figure of merit and its components \citep{pontius2008comparing, varga2019intensity}, that exposes this. Without them, the unweighted networks, which simulate only about half of the observed change, would look like the best models. The same issue affects model selection. A validation criterion such as macro F1 over all cells rewards persistence, and choosing models by FoM on the validation blocks would suit the purpose of a land-change model better.

\subsection{Attention gates and drivers}
Attention gates are usually justified by their ability to focus on the relevant parts of an image. Here they added 87,593 parameters (1.0\%) and gave higher coefficients to changing cells, but they did not make the simulation more accurate than a plain U-Net CA. A $32 \times 32$ convolutional neighbourhood seems to give enough context on its own for the patterns in these maps. We report AttentionUNet-CA because it was the model we set out to test, but a U-Net CA does at least as well and is simpler.

Among the drivers, population density and distance to the CBD carry most of the information; both describe how close a cell is to the existing city. Because GHS-POP places people on built-up land, the density layer partly restates the built-up pattern at the start of each decade. Road distance adds little, perhaps because the $32 \times 32$ window already shows the ribbons of built-up land that line the main roads.

\subsection{Limitations}
\label{sec:limitations}
The study has several limitations. GLC\_FCS30D is about 81\% accurate overall and under-represents vegetation in this area (Section~\ref{sec:check}), and we did not carry out our own accuracy assessment with field or high-resolution reference data. Resampling the 30~m maps to about 50~m by majority removes some small features. Apart from population density, the drivers are static. The road network comes from current OpenStreetMap data, so it may include roads built during the test decade. That could favour the models slightly, although road distance had little effect on the results. The Copernicus DEM is a surface model that includes buildings and trees, but elevation played almost no part in the simulations. Three seeds per configuration is a small number, and the three seeds of the 2030 projection differ considerably in the amount of new built-up land. We studied a single city. Finally, the models predict a decade ahead in one step, and the 2030 projection assumes that the transitions of 1990--2020 continue.

\section{Conclusions}
\label{sec:conclusion}
We tested a convolutional cellular automaton with attention gates, AttentionUNet-CA, on Lucknow for 1990--2020. The land-cover series was consistent over time, the test decade was kept out of training and model selection, the population layer never postdated the start of the decade being simulated, every configuration was run with three seeds, and the model was compared with a null model, a random-forest CA and a U-Net CA without attention.

Only 8.1\% of the study area changed in the test decade, so no model reached the overall agreement of the null model (91.92\%). On change itself, the class-weighted U-Net CA had the highest figure of merit (22.0\%), followed by the class-weighted AttentionUNet-CA (21.0\%) and RF-CA (20.3\%). These margins were comparable to the variation between seeds, and the networks estimated the amount of change more accurately than RF-CA. Class weighting raised FoM by 1--2 points by making the networks simulate more change, at the cost of more false alarms. The attention gates did not improve accuracy over a plain U-Net CA. Population density and distance to the CBD were the most influential drivers. In the projection, built-up land grows by 22--35\% between 2020 and 2030, mostly at the edge of the city.

For deep-learning land-change models, these results suggest that evaluation against a null model with change-based measures matters more than refinements to the network such as attention gates. Next steps are to select models with a change-based criterion, to use road networks for each date, to assess the land-cover maps against independent reference data, and to repeat the study in other cities.

\section*{Data availability}
All data used in this study are openly available: GLC\_FCS30D land cover \citep{zhang2024glcfcs30d}, the Copernicus DEM \citep{esa2022copdem}, OpenStreetMap \citep{osm2026} and GHS-POP \citep{schiavina2023ghspop}, and, for the independent check, GHS-BUILT-S \citep{pesaresi2023ghsbuilt} and ESA WorldCover \citep{zanaga2022worldcover}.

\bibliographystyle{elsarticle-harv}
\bibliography{references}

\end{document}